\documentclass{article}

\usepackage{PRIMEarxiv}

\usepackage[utf8]{inputenc} 
\usepackage[T1]{fontenc}    
\usepackage{hyperref}       
\usepackage{url}            
\usepackage{booktabs}       
\usepackage{amsfonts}       
\usepackage{nicefrac}       
\usepackage{microtype}      
\usepackage{lipsum}
\usepackage{tablefootnote}
\usepackage{color, colortbl}
\usepackage[misc]{ifsym} 
\usepackage{soul}
\usepackage{amsmath}
\usepackage{fixltx2e}
\usepackage{afterpage}
\usepackage{caption}
\usepackage{fancyhdr}       
\usepackage{graphicx}       
\graphicspath{{media/}}     
\usepackage[sorting=none, style=numeric-comp]{biblatex} 
\definecolor{brown}{RGB}{229,225,224}
\usepackage{float}

\usepackage{graphicx}
\usepackage[flushleft]{threeparttable}

\title{Beam Geometry and Input Dimensionality: Impact on Sparse-Sampling Artifact Correction for Clinical CT with U-Nets
}

\author{
  Tina Dorosti$^{1-3}$ \Letter \hspace{0.3em}, Johannes Thalhammer$^\mathrm{{1-4}}$, Sebastian Peterhansl$^\mathrm{{1, 2}}$, Daniela Pfeiffer$^\mathrm{{3, 4}}$,\\
  \textbf{Franz Pfeiffer$^\mathrm{{1-4}}$, Florian Schaff$^\mathrm{{1, 2}}$} \\\\
  1 Chair of Biomedical Physics, Department of Physics, School of Natural Sciences\\
  2 Munich Institute of Biomedical Engineering\\
  3 Institute for Diagnostic and Interventional Radiology, School of Medicine and Health, TUM Klinikum\\
  4 Institute for Advanced Study\\\\
  Technical University of Munich, Germany\\
  \Letter \hspace{0.3em} \texttt{tina.dorosti@tum.de} \\
}

\begin{document}
\maketitle

\section*{Key points:}
\begin{enumerate}
  \item Effects of beam geometry: 3D and 2.5D data don’t improve removing beam-geometry-dependant artifacts.
  \item Directionality comparison: 2D axial patches work best for artifact reduction, and coronal patches are worst.
  \item Neighboring slices as 2.5D: No improvement over the patch-wise 2.5D method as information bleeds in from neighboring slices.
\\
\end{enumerate}

\textbf{Abbreviations:} Computed tomography (CT), Mean squared error (MSE), Structural similarity index measure (SSIM)


\newpage
\begin{abstract}
This study aims to investigate the effect of various beam geometries and dimensions of input data on the sparse-sampling streak artifact correction task with U-Nets for clinical CT scans as a means of incorporating the volumetric context into artifact reduction tasks to improve model performance. A total of 22 subjects were retrospectively selected (01.2016-12.2018) from the Technical University of Munich's research hospital, TUM Klinikum rechts der Isar. Sparsely-sampled CT volumes were simulated with the Astra toolbox for parallel, fan, and cone beam geometries. 2048 views were taken as full-view scans. 2D and 3D U-Nets were trained and validated on 14, and tested on 8 subjects, respectively. For the dimensionality study, in addition to the 512x512 2D CT images, the CT scans were further pre-processed to generate a so-called '2.5D', and 3D data: Each CT volume was divided into 64x64x64 voxel blocks. The 3D data refers to individual 64-voxel blocks. An axial, coronal, and sagittal cut through the center of each block resulted in three 64x64 2D patches that were rearranged as a single 64x64x3 image, proposed as
2.5D data. Model performance was assessed with the mean squared error (MSE) and structural similarity index measure (SSIM). For all geometries, the 2D U-Net
trained on axial 2D slices results in the best MSE and SSIM values, outperforming the 2.5D and 3D input data dimensions.\\\\


\end{abstract}
\section*{Introduction}
Computed tomography (CT) is a popular modality for non-invasive three-dimensional (3D) imaging in clinics. Yet, it comes at the cost of exposing patients to ionization radiation, which should be minimized to doses as low as reasonably achievable \cite{yeung_alara_2019}. One method for reducing the exposure to X-ray radiation in a CT scan is by sparsely sampling the acquired projection views. This sparse sampling results in streaky artifacts, hindering the visibility of detailed and fine structures in the final reconstructed CT volume. The fewer the sampled views used for reconstruction, the more exaggerated the resulting artifacts are \cite{ries_improving_2024, thalhammer_improving_2024}. 

Recent developments in machine intelligence have allowed for various deep-learning-based models to successfully remove such streak artifacts in the post-processing of CT data. Commonly, streak artifacts are corrected for with deep-learning models on a 2D scheme, in which all 2D axial slices of the CT volume are processed individually \cite{ries_improving_2024,thalhammer_improving_2024, cheslerean-boghiu_wnet_2023, wu_drone_2021}. Given the volumetric nature of CT scans, this method disregards the 3D relation of the streaks during the artifact removal process. Specifically, for CT beam geometries other than parallel beam, such as cone beam geometry, the streaks depend on the location of the axial slice within the volume. 3D deep learning models could address this issue; however, the computational cost associated with the volumetric data drastically limits 3D models' complexity, size, and performance \cite{wang_multi-view_2021, cicek_3d_2016}.   

To better incorporate 3D information in the deep-learning pipeline, some works have reported improved performance by taking multiple 2D axial slices as input to the  model for reconstruction, classification, or detection tasks \cite{poon_2hDMC_detect_2023, zeng_2hdMC_detect_2023, takao_2hDMC_detect_2022, ziabari_2hDMC_reconstruction_2018}. Furthermore, so-called 2.5D approaches have been suggested for segmentation \cite{yoo_2hDMV_segmentation_2023, song_2hDMV_segmentation_2021, zheng_2hDMV_segmentation_2020} and classification tasks \cite{zhang_2hDMV_detect_2024, geng_2hdMV_detect_2019} by providing deep-learning models with 2D information from axial, coronal, and sagittal views, fused or assembled to represent the 3D structure. In doing so, the high computation cost of directly utilizing 3D input is bypassed while the model still benefits from the necessary volumetric information \cite{zhang_2hDMV_detect_2024, yoo_2hDMV_segmentation_2023, song_2hDMV_segmentation_2021, wang_multi-view_2021, zheng_2hDMV_segmentation_2020, geng_2hdMV_detect_2019}.


In this study, we aim to investigate the effect of various beam geometries and input dimensionalities on the sparse-sampling streak artifact correction with U-Nets for clinical CT scans. More specifically, comparing 2D and 3D input data to the proposed 2.5D pre-processing approach. In this 2.5D pre-processing, a three-channel image is generated from three orthogonal slices through the center of a 3D volume. U-Net models were trained and tested individually for data from parallel, fan, and cone geometries to assess the influence of the beam geometry on the streak artifact correction task. Lastly, an ablation study compared the effect of  axial, coronal, and sagittal images for the 2D U-Net model. 

\section*{Methods}
\subsection*{Dataset}

A total of 22 CT scans from 22 different subjects were retrospectively selected (01.2016-12.2018) from the picture archiving and communication system of the Technical University of Munich's (TUM) research hospital, TUM Klinikum. Approval from the institutional review board was received, and the requirement for written informed consent was waived as data was analyzed anonymously and retrospectively. Lung metastasis was observed in all subjects, with no other lung diseases reported. The subject selection process was as follows: Initially, a total of 30 subjects were considered. Next, perihilar localization of metastases eliminated four subjects. Lastly, cases with atelectasis, additional pleural effusion, or other lung diseases were excluded, resulting in the final sample size of 22 subjects, corresponding to a total of 11016 axial CT slices of 512x512 pixels. Data was split randomly on a subject level with a train, validation, test ratio of approximately 50:10:40. Subject demographics are reported in \autoref{tab:demographics}.\\

\begin{table}[h!]
\centering
\begin{threeparttable}
\caption{\small Subject demographics (\textit{n} = 22)}
\label{tab:demographics}
\small
  \centering
    \begin{tabular}{lcccc}
    \toprule 
     &  Train  &  Validation & Test \\
    \midrule
    \midrule
    Male & 5 & 2 & 4 \\
    Female & 7 & 0 & 4 \\
    Age (years) & 65.8 $\pm$ 11.6 & 77.0 $\pm$ 4.00 & 61.9 $\pm$ 11.8 \\
    CT slice count & 6864 & 1092 & 3060\\
    \bottomrule
    \end{tabular}
    \begin{tablenotes}
    \small
      \item Age is given as mean $\pm$ standard deviation.
    \end{tablenotes}
    \end{threeparttable}
\end{table}

\subsection*{Data Preparation}
The Astra toolbox (version 2.1.1) was utilized to generate simulated sparse CT images \cite{van_aarle_astra_2015, van_aarle_fast_2016, palenstijn_astra_2011}: CT scans were forward projected to 2048-views singorams, and subsequently undersampled to simulate different levels of sparse-view CT via the filtered back projection algorithm for the parallel and fan beam geometries, and with the Feldkamp-Davis-Kress algorithm for the cone beam geometry. The full-view was reconstructed from all 2048 views; the sparsely sampled CT scans used subsets of 32, 64, and 128 views, respectively. Intensity values were clipped to a wide Hounsfield Unit (HU) window range (width = 2048, level = 0) HU and subsequently normalized to values between zero and one. All CT images were reconstructed onto a 512x512 pixels image matrix in the axial plane. The ground-truth data was taken as the difference between full- and sparse-view scans for each level of subsampling. Artifact-corrected CT images were then obtained by subtracting the U-Net artifact prediction from the sparse-view input data.

\begin{figure*}[h!]
\centerline{\includegraphics[width=\textwidth, trim={0 9cm 0 9cm}, clip]{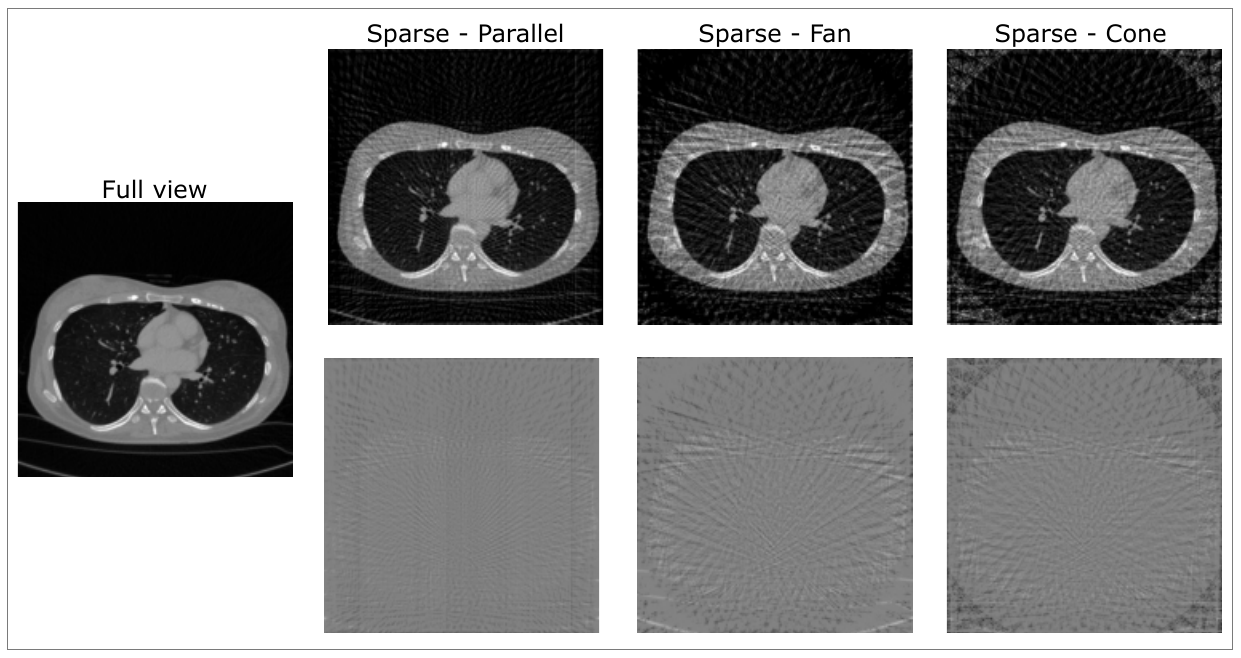}}
\caption{\small Example of a 2D image reconstructed with 64 views for parallel, fan, and cone beam geometries, in comparison to the same slice reconstructed with 2048 views, taken as the full view. Normalized CT images range from 0 to 1. The bottom row shows the difference between the full view and the sparse images provided to the model as the ground truth label. Difference images range from -1 to 1. }
\label{fig:beamGeometry_exampleSlices}
\end{figure*}

For all beam geometries, a source-to-object distance of 570 mm and a source-to-detector distance of 1040 mm were used in accordance with the geometry of the clinical CT systems as reported in the corresponding DICOM files. \autoref{fig:beamGeometry_exampleSlices} depicts an example axial slice sparsely-sampled with 64 views for all beam geometries, in comparison to the full-view reconstruction. 

For the dimensionality study, in addition to the conventional 2D axial CT slices as input, the CT scans were further pre-processed to generate the following data formats, namely 3D, 2.5D, and 2D3ch data. \autoref{fig:dataPreProc+Nets}A demonstrates the data pre-processing schemes. For the fully 3D data, due to compute limitations, the CT volumes were divided into smaller blocks. The volume was split into 64x64x64 blocks such that each block included 48 unique voxels and neighboring blocks had an overlap of 8 voxels from each side. Zero-padding was applied when necessary to ensure each CT volume was divisible into an integer number of 64x64x64 blocks. The 3D data, therefore, refers to individual 64x64x64 blocks. The 2.5D data refers to an axial, coronal, and sagittal cut through the center of each 3D block, resulting in three 64x64 2D image patches rearranged as a single 64x64x3 image. Due to data limitations, the 2.5D approach was only applied on the block level and not the CT volume level. The 2D 64x64 image patches were only considered for the ablation study. Use of larger 2.5D data (i.e. 128 voxels) as well as a narrower HU window setting was analyzed for the train and validation sets, as shown in \autoref{fig:loss_window_blockSize}, and the combination with the best train and validation MSE values were pursued further, namely the wide window setting and the 64x64x64 block data, as specified in this section. Lastly, the 2D3ch data format was created by stacking three neighboring axial slices, leading to an input size of 512x512x3.

\subsection*{Network Architecture and Training}
\begin{figure*}[h!]
\centerline{\includegraphics[width=\textwidth, trim={0 2.5cm 0 2.5cm}, clip]{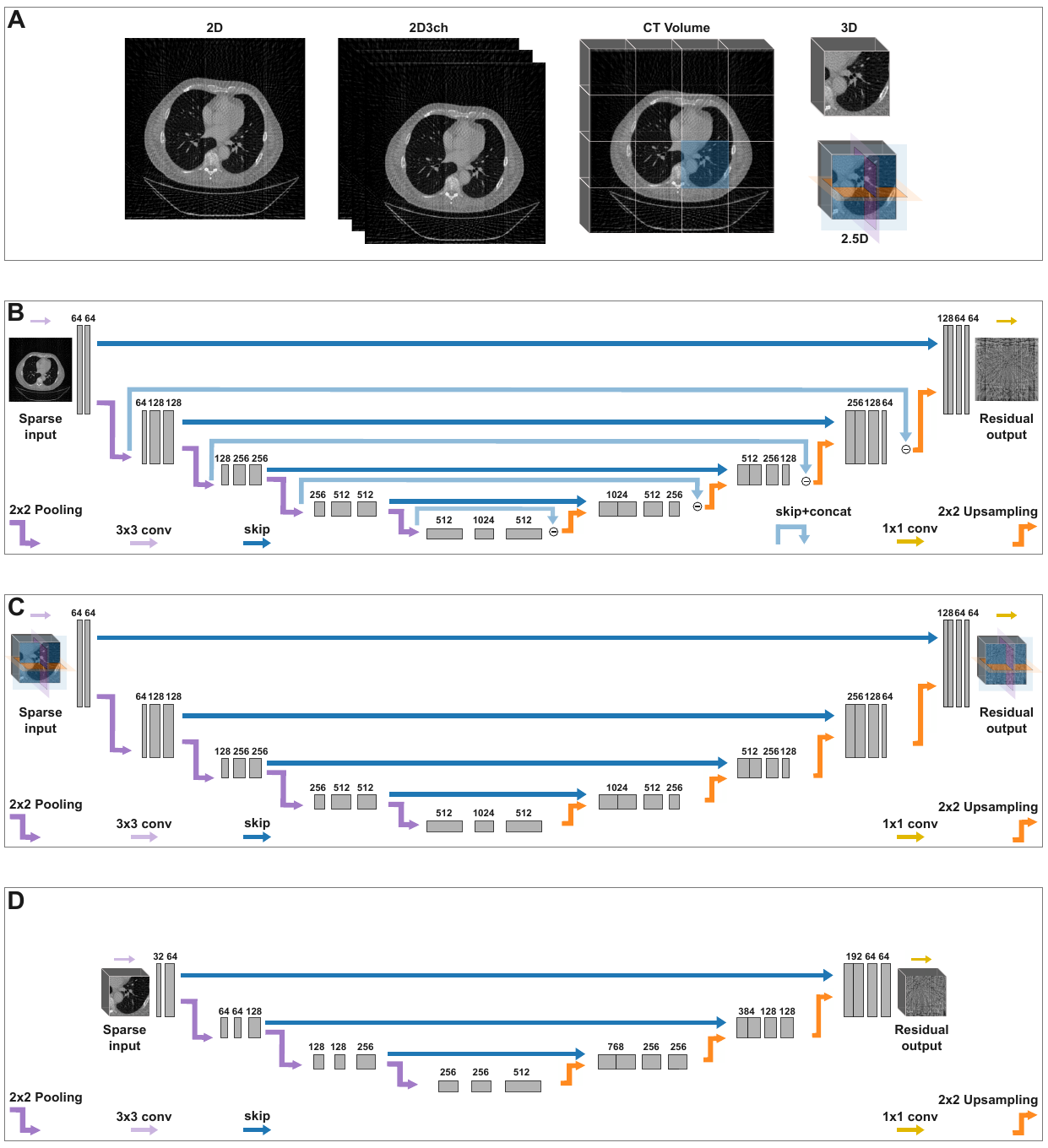}}
\caption{\small \textbf{A)} Data preprocessing for dimensionality comparison between two-dimensional (2D) axial images of size 512x512 pixels, two-dimensional images and their corresponding direct neighboring images in the CT volume (2D3ch), three-dimensional (3D) blocks of 64x64x64, and two-and-a-half dimensional (2.5D) data corresponding to three orthogonal cuts of 64x64 2D patches through the 3D block, \textbf{B)} the architecture of the dual-frame U-Net utilized for training on 2D data \cite{han_framing_2018}, \textbf{C)} the architecture of the U-Net utilized for training on 2.5D data and the corresponding 2D patches for the ablation study \cite{ronneberger_u-net_2015}, \textbf{D)} the architecture of the 3D U-Net utilized for training on 3D blocks \cite{cicek_3d_2016}.}
\label{fig:dataPreProc+Nets}
\end{figure*}

Based on previous work of sparse-view artifact correction for 2D CT images \cite{ries_improving_2024}, the dual-frame U-Net variant as depicted in \autoref{fig:dataPreProc+Nets}B was utilized for the 2D data \cite{han_framing_2018}. For the 2D3ch and 2.5D data, the vanilla U-Net \cite{ronneberger_u-net_2015} shown in \autoref{fig:dataPreProc+Nets}C was implemented as it was found to outperform the dual-frame variant for the 2.5D data reconstructed with parallel beam data, based on the results in \autoref{fig:loss_unet_comparison}. A 3D U-Net variant, as illustrated in \autoref{fig:dataPreProc+Nets}D, was respectively used for the 3D data \cite{cicek_3d_2016}. NVIDIA RTX A4000 (16 GB VRAM) and RTX 3090 (24 GB VRAM) graphics cards were utilized to train the models for the 2D, 2D3ch, and 2.5D data. The 3D U-Net model was trained with an NVIDIA A100 (80 GB VRAM) graphics card. All models were implemented with the Keras API of the Tensorflow library (version 2.4.0) with random initialization \cite{keras, tensorflow}. For each level of subsampling, a new model was trained from scratch with a mean squared error (MSE) loss and the ADAM optimizer. The learning rate decayed exponentially per epoch $n$ with $lr_n = lr_{n-1} \cdot e^{-0.1}$, after an initial value of $lr = 0.001$. Early stopping was implemented with patience of 20 epochs, and the model weights from the epoch with the smallest validation loss were taken for inference on the test set. The 2D model was trained with a batch size of six for a total of 30 epochs. The 2.5D model was trained with a batch size of 16 for a total of 30 epochs. The 3D model was trained with a batch size of 16 and a total of 20 epochs. As the training process exhibited no trend of overfitting, no data augmentation was applied.

\subsection*{Evaluation Metrics}
Model performance was evaluated with the MSE and the structural similarity index measure (SSIM), using the scikit image library (version 0.18.3) \cite{scikit-image}.

\section*{Results}

Here, the results for spare-view CT artifact correction of the test set ($n$ = 8) are presented as follows: First, the impact of the 2D, 2.5D, and 3D data given various beam geometries for all sparse-view angles is reported. Next, an ablation study explores the impact of the patch size for the 2.5D data, given the parallel beam geometry. Last, for the sparse data with 128 views, the 2D and multi-channel 2D3ch data given various beam geometries are compared.

\subsection*{Effects of Beam Geometry and Data Dimensionality }\label{subsec:results_geomComp}

\begin{figure*}[h!]
\centerline{\includegraphics[width=\textwidth, trim={0 7cm 0 7cm}, clip]{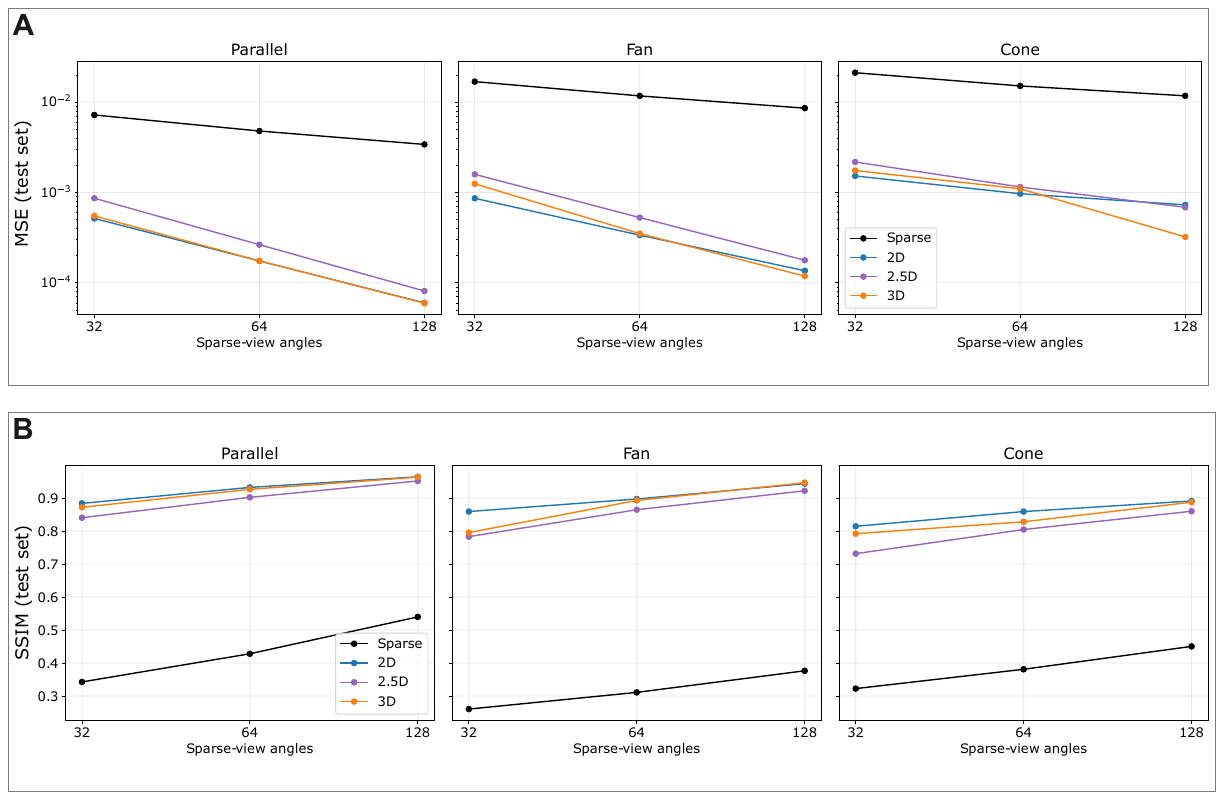}}
\caption{\small The mean \textbf{A)} MSE and \textbf{B)} SSIM values for parallel, fan, and cone beam sparse and artifact-corrected test data for models trained with 2D, 2.5D, and 3D data.}
\label{fig:geomComp_MSE_SSIM}
\end{figure*}

The results of the U-Nets trained and tested on 2D, 2.5D, and 3D data were compared for parallel, fan, and cone beam geometries to examine whether sparse-view artifacts were removed more effectively by providing volumetric information about the data. The mean MSE and SSIM values on the sparse and predicted data for 32, 64, and 128 views are provided in \autoref{fig:geomComp_MSE_SSIM}, both of which improve with increasing number of views as the number of streaks decreases.
\autoref{tab:geomComp_MSE_SSIM_supp} reports the exact MSE and SSIM values shown in \autoref{fig:geomComp_MSE_SSIM}. 

\begin{figure*}[h!]
\centerline{\includegraphics[width=0.98\textwidth, trim={0 0 0 2cm}, clip]{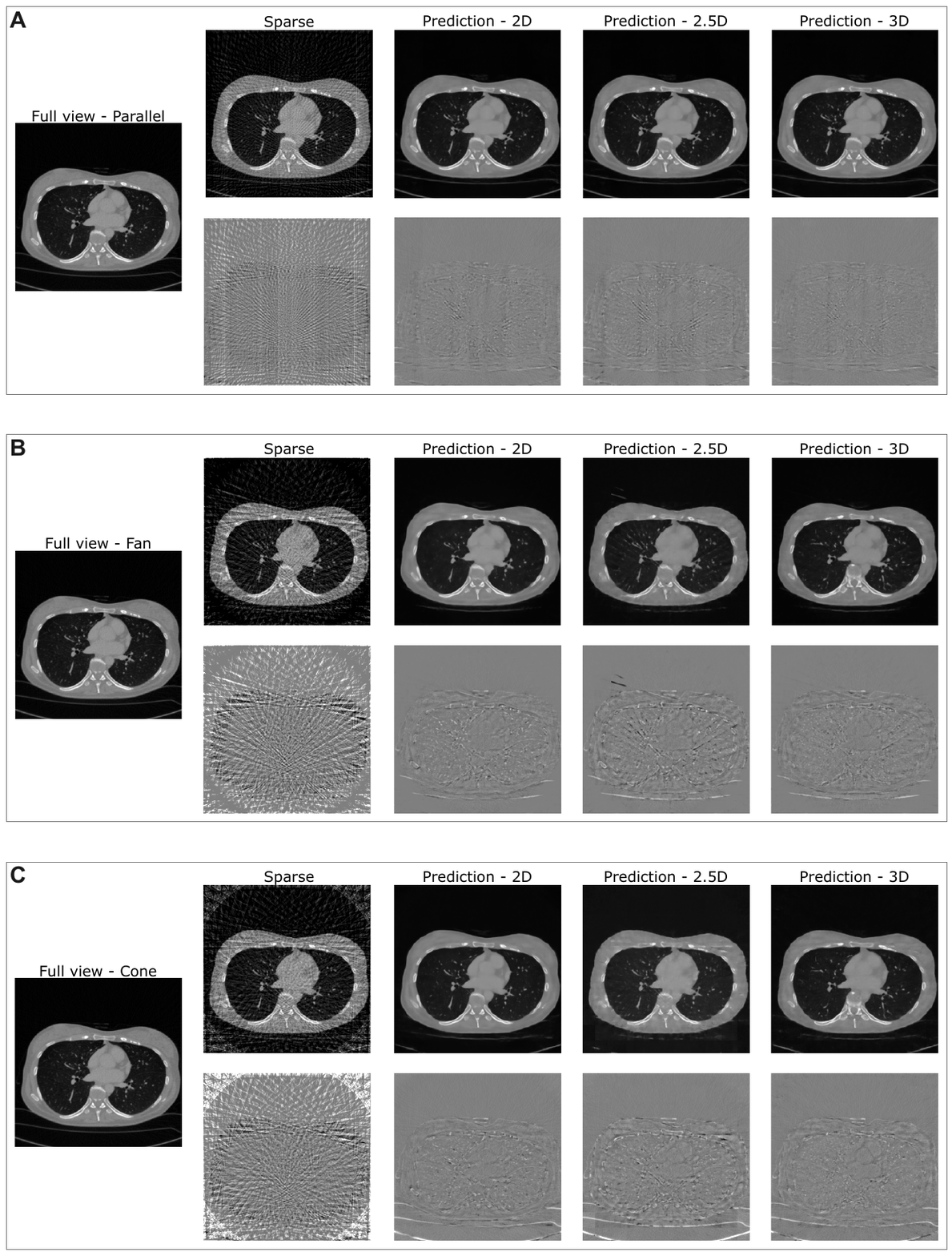}}
\caption{\small Example of an axial image reconstructed with 64 views for \textbf{A)} parallel, \textbf{B)} fan, and \textbf{C)} cone beam geometries for 2D, 2.5D, and 3D U-Net variants, in comparison to the same slice reconstructed with 2048 views, taken as the full-view image. Normalized CT images range from 0 to 1. The bottom row shows the difference between each image and the full view. Difference images range from -0.3 to 0.3.}
\label{fig:geomComp_exampleSlices_axial_64}
\end{figure*}

\begin{figure*}[h!]
\centerline{\includegraphics[width=\textwidth, trim={0 8cm 0 8cm}, clip]{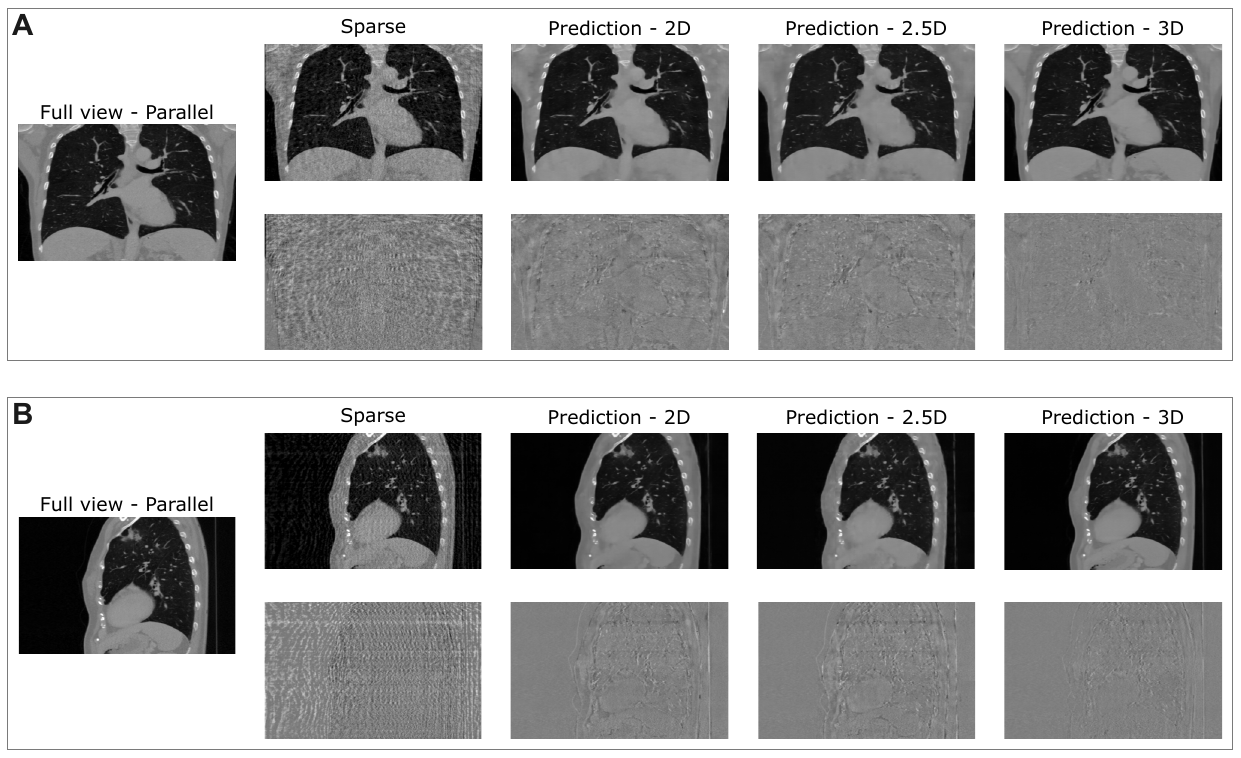}}
\caption{\small Example of a \textbf{A)} coronal and \textbf{B)} sagittal image reconstructed with 64 views for parallel beam geometry for 2D, 2.5D, and 3D U-Net variants, in comparison to the same slice reconstructed with 2048 views, taken as the full-view images. Normalized CT images range from 0 to 1. The bottom row shows the difference between each image and the full view. Difference images range from -0.3 to 0.3.}
\label{fig:dimComp_exampleSlices_cor_sag_64}
\end{figure*}

Overall, compared to the fan and cone beam geometries, the parallel beam geometry results in lower MSE and higher SSIM values than the fan beam, followed by the cone beam. In the case of the parallel beam, the 2D and 3D methods are on par and outperform the 2.5D methods for both metrics. For fan beam geometry, the 3D and 2D approaches are on par for 64 and 128 views in terms of the SSIM, and the 3D method is slightly better than 2D for 128 views in terms of the MSE. In the case of the cone beam geometry, the 3D data improves the MSE for 128-views and is on par with the 2D method, given the SSIM metric. Example axial slices for 64 views, given all dimensionality and beam geometry comparison are provided in \autoref{fig:geomComp_exampleSlices_axial_64}. Example slices of coronal and sagittal slices given the parallel beam geometry, processed by U-Nets for all data dimensionalities, are presented in \autoref{fig:dimComp_exampleSlices_cor_sag_64}

\subsection*{Ablation Study}

\begin{figure*}[h!]
\centerline{\includegraphics[width=\textwidth, trim={0 11cm 0 11cm}, clip]{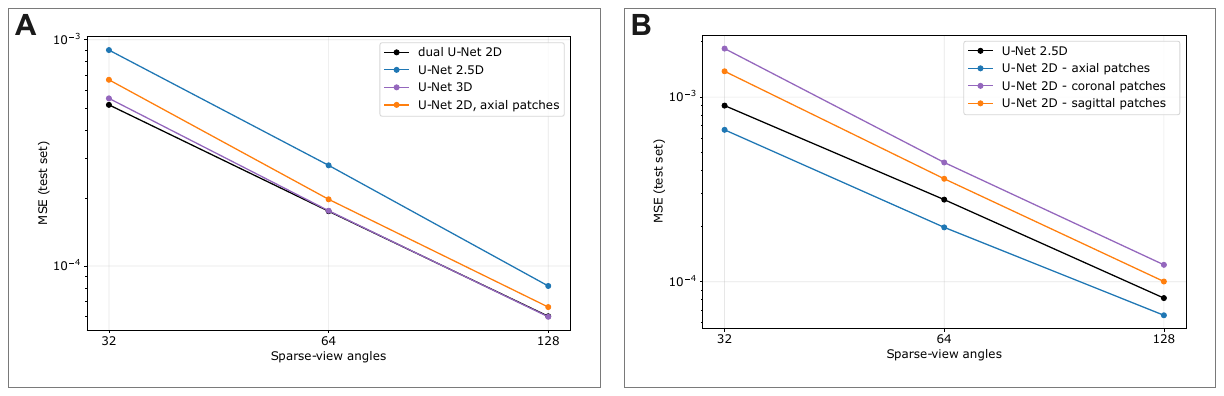}}
\caption{\small The mean squared error (MSE) values for all parallel-beam artifact-corrected test data \textbf{A) }for models trained with 2D, 2.5D, 3D, and 2D axial patch input data over all sparse-view angles and, \textbf{B) } for the U-Nets trained individually on 2.5D, and the corresponding 2D patches from the axial, coronal, and sagittal directions.}\label{fig:dimComp_ablation}
\end{figure*}

To evaluate the influence of the direction of the 2D patches on the 2.5D data, an ablation study was conducted by training three individual 2D U-Nets from scratch on 64x64 patches, each corresponding to the axial, coronal, and sagittal components of the 2.5D channels, respectively. The data was reconstructed using parallel beam geometry. \autoref{fig:dimComp_ablation} shows that the model with 2D axial patches as input data results in the best MSE values, followed by 2.5D data, sagittal patches, and lastly, coronal patches. The model with 2D full-sized axial slices outperforms the model with 2D axial patches. 

\subsection*{Multi-channel slices as a 2.5D approach}
\begin{figure*}[h!]
\centerline{\includegraphics[width=0.98\textwidth, trim={0 8cm 0 8cm}, clip]{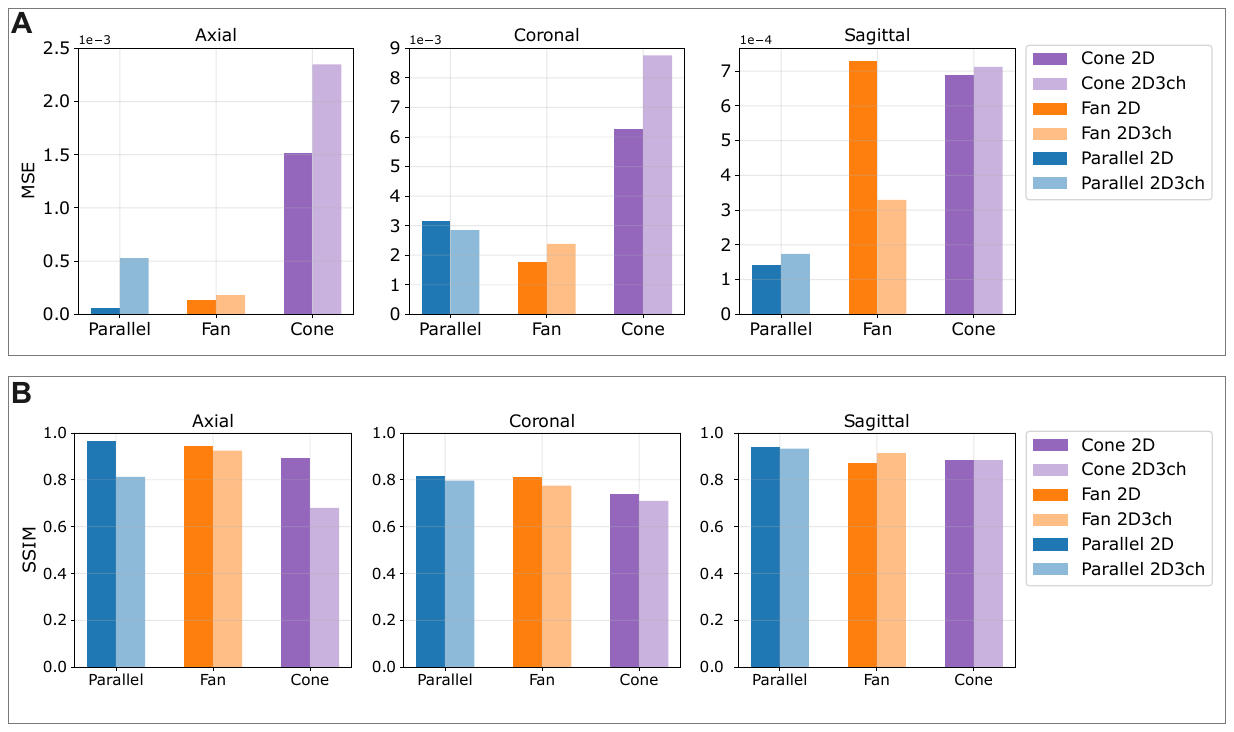}}
\caption{\small Comparison of the mean \textbf{A)} MSE and \textbf{B)} SSIM values for 2D and 2D3ch 128-sparse-view data for models trained and tested on axial, coronal, and sagittal views given the parallel, fan, and cone beam geometries.}
\label{fig:geomComp_2d3ch}
\end{figure*}

The 2D3ch is commonly referred to and utilized as a 2.5D approach  in literature \cite{poon_2hDMC_detect_2023, zeng_2hdMC_detect_2023, takao_2hDMC_detect_2022, ziabari_2hDMC_reconstruction_2018}. Therefore, the effect of beam geometry on 128-view 2D3ch data was additionally explored. The comparison of the MSE and SSIM values for 2D3ch and 2D data is provided in \autoref{fig:geomComp_2d3ch}. For the axial view, the corresponding values are reported in \autoref{tab:geomComp_MSE_SSIM_128}. The 2D method outperformed the 2D3ch method for the axial view regardless of the beam geometry. For the coronal and sagittal views, inconsistencies were observed between the mean MSE and SSIM values, as well as the comparison between the 2D and the 2D3ch approach.

\section*{Discussion}

In this retrospective study, the effect of the dimensionality of the input data on U-Nets was explored, specifically in relation to the simulated reconstruction beam geometry. Additionally, for comparison to similar work in the literature, adjacent neighboring slices (2D3ch) as a 2.5D approach were compared to the proposed 2.5D approach.

For parallel beam geometry, the 2D model outperformed 2.5D and 3D models, considering the MSE and the SSIM metrics, as well as computational effort. This can be understood as the parallel beam geometry did not cause a strong influence on the pattern of the streaks based on the location of the axial slice. The ablation study implied that providing 2D axial patches to the model results in the best performance in comparison to sagittal, followed by coronal 2D patches. Given the higher resolution of axial slices and the fact that the pattern of streaks was more homogenous in this view, the better performance of the 2D axial patches could be attributed to the easier problem the model had to correct for. Additionally, there are fewer slices containing only air in axial, sagittal, and coronal planes, respectively, aligning with the observation from the ablation study. Notably, 2D axial slices showed a higher performance than the 2D axial patches. These observations pointed to the weaker performance of the 2.5D model in comparison to the 2D model.

When looking at the fan and cone beam geometries, the 2D and 3D model performances were on par, indicating the 2D model sufficiently captured the necessary volumetric streaking structure. Based on observations of the ablation study, it can be speculated that the 2.5D model's performance was set back due to the limited input patch size, as well as the extreme variability of streak patterns in the sagittal and coronal view patches provided as additional channels. The block size could also be a contributing factor for why the 3D model did not outperform the 2D model despite the additional volumetric information provided to the model. The trade-off between input size and the volumetric information likely prevented the 3D model from outperforming the 2D model.

In this study, the addition of neighboring slices as a means of providing volumetric information was not beneficial to the streak correction in comaprison to the model trained on 2D single-channel data for 128 views. This is contrary to the reports in the literature on the benefits of 2D3ch data for reconstruction, classification, or detection tasks \cite{poon_2hDMC_detect_2023, zeng_2hdMC_detect_2023, takao_2hDMC_detect_2022, ziabari_2hDMC_reconstruction_2018}. The variability of streak patterns in the neighboring slices could be a potential explanation for this behavior. Nonetheless, a more detailed investigation of the number of neighboring slices needed, as well as the influence of the severity of streaking provided by fewer or more sparse-view angles, is needed.

The main limitation of this work is the small sample of data points obtained from one medical center. Additionally, the proposed 2.5D and 3D models were not able to capture the essential volumetric information to benefit the streak artifact correction, possibly due to the limitations on the input size of the patches and blocks. Future work can focus on hybrid 2D, 3D methods to bypass the computational limitations while allowing for inter-slice consistency of volumetric data.

A review of dimensionality comaprison of 2D, 2.5D, and 3D input data generated with parallel, fan, and cone beam geometries to correct for sparse CT streak artifacts with U-Nets demonstrated that for all geometries, the 2D U-Net trained on axial 2D slices results in the best MSE and SSIM values.

\newpage
\printbibliography 

@article{yeung_alara_2019,
	title = {The “As Low As Reasonably Achievable” ({ALARA}) principle: a brief historical overview and a bibliometric analysis of the most cited publications},
	volume = {54},
	rights = {© {EDP} Sciences 2019},
	issn = {0033-8451, 1769-700X},
	url = {https://www.radioprotection.org/articles/radiopro/abs/2019/02/radiopro190010/radiopro190010.html},
	doi = {10.1051/radiopro/2019016},
	shorttitle = {The “As Low As Reasonably Achievable” ({ALARA}) principle},
	pages = {103--109},
	number = {2},
	journaltitle = {Radioprotection},
	shortjournal = {Radioprotection},
	author = {Yeung, A. W. K.},
	urldate = {2025-04-28},
	date = {2019-04-01},
	langid = {english},
	note = {Number: 2
Publisher: {EDP} Sciences},
}

@misc{cicek_3d_2016,
	title = {{3D} {U}-{Net}: {Learning} {Dense} {Volumetric} {Segmentation} from {Sparse} {Annotation}},
	shorttitle = {{3D} {U}-{Net}},
	url = {http://arxiv.org/abs/1606.06650},
	doi = {10.48550/arXiv.1606.06650},
	urldate = {2025-04-28},
	publisher = {arXiv},
	author = {Çiçek, Özgün and Abdulkadir, Ahmed and Lienkamp, Soeren S. and Brox, Thomas and Ronneberger, Olaf},
	month = jun,
	year = {2016},
	note = {arXiv:1606.06650 [cs]},
}

@article{ries_improving_2024,
	title = {Improving image quality of sparse-view lung tumor {CT} images with {U}-{Net}},
	volume = {8},
	issn = {2509-9280},
	url = {https://doi.org/10.1186/s41747-024-00450-4},
	doi = {10.1186/s41747-024-00450-4},
	number = {1},
	urldate = {2025-04-28},
	journal = {European Radiology Experimental},
	author = {Ries, Annika and Dorosti, Tina and Thalhammer, Johannes and Sasse, Daniel and Sauter, Andreas and Meurer, Felix and Benne, Ashley and Lasser, Tobias and Pfeiffer, Franz and Schaff, Florian and Pfeiffer, Daniela},
	month = may,
	year = {2024},
	pages = {54},
}

@article{thalhammer_improving_2024,
	title = {Improving {Automated} {Hemorrhage} {Detection} at {Sparse}-{View} {CT} via {U}-{Net}–based {Artifact} {Reduction}},
	volume = {6},
	url = {https://pubs.rsna.org/doi/abs/10.1148/ryai.230275},
	doi = {10.1148/ryai.230275},
	number = {4},
	urldate = {2025-04-28},
	journal = {Radiology: Artificial Intelligence},
	author = {Thalhammer, Johannes and Schultheiß, Manuel and Dorosti, Tina and Lasser, Tobias and Pfeiffer, Franz and Pfeiffer, Daniela and Schaff, Florian},
	month = jul,
	year = {2024},
	note = {Publisher: Radiological Society of North America},
	pages = {e230275},
}

@article{zheng_2hDMV_segmentation_2020,
	title = {Improving the slice interaction of 2.5D {CNN} for automatic pancreas segmentation},
	volume = {47},
	issn = {2473-4209},
	url = {https://onlinelibrary.wiley.com/doi/abs/10.1002/mp.14303},
	doi = {10.1002/mp.14303},
	pages = {5543--5554},
	number = {11},
	journaltitle = {Medical Physics},
	author = {Zheng, Hao and Qian, Lijun and Qin, Yulei and Gu, Yun and Yang, Jie},
	urldate = {2025-04-30},
	date = {2020},
	langid = {english},
	note = {\_eprint: https://onlinelibrary.wiley.com/doi/pdf/10.1002/mp.14303},
}

@article{yoo_2hDMV_segmentation_2023,
	title = {Comparison of 2D, 2.5D, and 3D segmentation networks for maxillary sinuses and lesions in {CBCT} images},
	volume = {23},
	issn = {1472-6831},
	url = {https://doi.org/10.1186/s12903-023-03607-6},
	doi = {10.1186/s12903-023-03607-6},
	pages = {866},
	number = {1},
	journaltitle = {{BMC} Oral Health},
	shortjournal = {{BMC} Oral Health},
	author = {Yoo, Yeon-Sun and Kim, {DaEl} and Yang, Su and Kang, Se-Ryong and Kim, Jo-Eun and Huh, Kyung-Hoe and Lee, Sam-Sun and Heo, Min-Suk and Yi, Won-Jin},
	urldate = {2025-04-30},
	date = {2023-11-15},
	langid = {english},
}

@article{song_2hDMV_segmentation_2021,
	title = {Bridging the {Gap} {Between} {2D} and {3D} {Contexts} in {CT} {Volume} for {Liver} and {Tumor} {Segmentation}},
	volume = {25},
	issn = {2168-2208},
	url = {https://ieeexplore.ieee.org/document/9416736},
	doi = {10.1109/JBHI.2021.3075752},
	number = {9},
	urldate = {2025-04-28},
	journal = {IEEE Journal of Biomedical and Health Informatics},
	author = {Song, Lei and Wang, Haoqian and Wang, Z. Jane},
	month = sep,
	year = {2021},
	pages = {3450--3459},
}

@article{cheslerean-boghiu_wnet_2023,
	title = {{WNet}: {A} {Data}-{Driven} {Dual}-{Domain} {Denoising} {Model} for {Sparse}-{View} {Computed} {Tomography} {With} a {Trainable} {Reconstruction} {Layer}},
	volume = {9},
	issn = {2333-9403},
	shorttitle = {{WNet}},
	url = {https://ieeexplore.ieee.org/document/10026634},
	doi = {10.1109/TCI.2023.3240078},
	urldate = {2025-04-28},
	journal = {IEEE Transactions on Computational Imaging},
	author = {Cheslerean-Boghiu, Theodor and Hofmann, Felix C. and Schultheiß, Manuel and Pfeiffer, Franz and Pfeiffer, Daniela and Lasser, Tobias},
	year = {2023},
	pages = {120--132},
}

@article{wu_drone_2021,
	title = {{DRONE}: Dual-Domain Residual-based Optimization {NEtwork} for Sparse-View {CT} Reconstruction},
	volume = {40},
	issn = {1558-254X},
	url = {https://ieeexplore.ieee.org/document/9424618},
	doi = {10.1109/TMI.2021.3078067},
	shorttitle = {{DRONE}},
	pages = {3002--3014},
	number = {11},
	journaltitle = {{IEEE} Transactions on Medical Imaging},
	author = {Wu, Weiwen and Hu, Dianlin and Niu, Chuang and Yu, Hengyong and Vardhanabhuti, Varut and Wang, Ge},
	urldate = {2025-04-30},
	date = {2021-11},
}

@inproceedings{wang_multi-view_2021,
	title = {Multi-view 3D Reconstruction with Transformers},
	url = {https://ieeexplore.ieee.org/document/9711318},
	doi = {10.1109/ICCV48922.2021.00567},
	eventtitle = {2021 {IEEE}/{CVF} International Conference on Computer Vision ({ICCV})},
	pages = {5702--5711},
	booktitle = {2021 {IEEE}/{CVF} International Conference on Computer Vision ({ICCV})},
	author = {Wang, Dan and Cui, Xinrui and Chen, Xun and Zou, Zhengxia and Shi, Tianyang and Salcudean, Septimiu and Wang, Z. Jane and Ward, Rabab},
	urldate = {2025-04-30},
	date = {2021-10},
	note = {{ISSN}: 2380-7504},
}

@misc{poon_2hDMC_detect_2023,
	title = {Detecting adrenal lesions on 3D {CT} scans using a 2.5D deep learning model},
	rights = {© 2023, Posted by Cold Spring Harbor Laboratory. This pre-print is available under a Creative Commons License (Attribution-{NoDerivs} 4.0 International), {CC} {BY}-{ND} 4.0, as described at http://creativecommons.org/licenses/by-nd/4.0/},
	url = {https://www.medrxiv.org/content/10.1101/2023.02.22.23286184v1},
	doi = {10.1101/2023.02.22.23286184},
	publisher = {{medRxiv}},
	author = {Poon, Sanson T. S. and Hanna, Fahmy W. F. and Lemarchand, François and George, Cherian and Clark, Alexander and Lea, Simon and Coleman, Charlie and Sollazzo, Giuseppe},
	urldate = {2025-04-30},
	date = {2023-02-24},
	langid = {english},
	note = {Pages: 2023.02.22.23286184},
}

@article{takao_2hDMC_detect_2022,
	title = {Deep-learning 2.5-dimensional single-shot detector improves the performance of automated detection of brain metastases on contrast-enhanced {CT}},
	volume = {64},
	issn = {1432-1920},
	url = {https://doi.org/10.1007/s00234-022-02902-3},
	doi = {10.1007/s00234-022-02902-3},
	pages = {1511--1518},
	number = {8},
	journaltitle = {Neuroradiology},
	shortjournal = {Neuroradiology},
	author = {Takao, Hidemasa and Amemiya, Shiori and Kato, Shimpei and Yamashita, Hiroshi and Sakamoto, Naoya and Abe, Osamu},
	urldate = {2025-04-30},
	date = {2022-08-01},
	langid = {english},
}

@article{zeng_2hdMC_detect_2023,
	title = {A 2.5D Deep Learning-Based Method for Drowning Diagnosis Using Post-Mortem Computed Tomography},
	volume = {27},
	issn = {2168-2208},
	url = {https://ieeexplore.ieee.org/abstract/document/9965594},
	doi = {10.1109/JBHI.2022.3225416},
	pages = {1026--1035},
	number = {2},
	journaltitle = {{IEEE} Journal of Biomedical and Health Informatics},
	author = {Zeng, Yuwen and Zhang, Xiaoyong and Kawasumi, Yusuke and Usui, Akihito and Ichiji, Kei and Funayama, Masato and Homma, Noriyasu},
	urldate = {2025-04-30},
	date = {2023-02},
}

@inproceedings{ziabari_2hDMC_reconstruction_2018,
	title = {2.5D Deep Learning For {CT} Image Reconstruction Using A Multi-{GPU} Implementation},
	url = {https://ieeexplore.ieee.org/abstract/document/8645364},
	doi = {10.1109/ACSSC.2018.8645364},
	eventtitle = {2018 52nd Asilomar Conference on Signals, Systems, and Computers},
	pages = {2044--2049},
	booktitle = {2018 52nd Asilomar Conference on Signals, Systems, and Computers},
	author = {Ziabari, Amirkoushyar and Ye, Dong Hye and Srivastava, Somesh and Sauer, Ken D. and Thibault, Jean-Baptiste and Bouman, Charles A.},
	urldate = {2025-04-30},
	date = {2018-10},
	note = {{ISSN}: 2576-2303},
}

@article{zhang_2hDMV_detect_2024,
	title = {Construction of a 2.5D Deep Learning Model for Predicting Early Postoperative Recurrence of Hepatocellular Carcinoma Using Multi-View and Multi-Phase {CT} Images},
	volume = {11},
	issn = {2253-5969},
	doi = {10.2147/JHC.S493478},
	pages = {2223--2239},
	journaltitle = {Journal of Hepatocellular Carcinoma},
	shortjournal = {J Hepatocell Carcinoma},
	author = {Zhang, Yu-Bo and Chen, Zhi-Qiang and Bu, Yang and Lei, Peng and Yang, Wei and Zhang, Wei},
	date = {2024},
	pmid = {39569409},
	pmcid = {PMC11577935},
}

@inproceedings{geng_2hdMV_detect_2019,
	title = {2.{5D} {CNN} model for detecting lung disease using weak supervision},
	volume = {10950},
	url = {https://www.spiedigitallibrary.org/conference-proceedings-of-spie/10950/109503O/25D-CNN-model-for-detecting-lung-disease-using-weak-supervision/10.1117/12.2513631.full},
	doi = {10.1117/12.2513631},
	urldate = {2025-04-28},
	booktitle = {Medical {Imaging} 2019: {Computer}-{Aided} {Diagnosis}},
	publisher = {SPIE},
	author = {Geng, Yue and Ren, Yinhao and Hou, Rui and Han, Songyue and Rubin, Geoffrey D. and Lo, Joseph Y.},
	month = mar,
	year = {2019},
	pages = {924--928},
}

@article{van_aarle_astra_2015,
	title = {The {ASTRA} {Toolbox}: {A} platform for advanced algorithm development in electron tomography},
	volume = {157},
	issn = {0304-3991},
	shorttitle = {The {ASTRA} {Toolbox}},
	url = {https://www.sciencedirect.com/science/article/pii/S0304399115001060},
	doi = {10.1016/j.ultramic.2015.05.002},
	urldate = {2025-04-28},
	journal = {Ultramicroscopy},
	author = {van Aarle, Wim and Palenstijn, Willem Jan and De Beenhouwer, Jan and Altantzis, Thomas and Bals, Sara and Batenburg, K. Joost and Sijbers, Jan},
	month = oct,
	year = {2015},
	pages = {35--47},
}

@article{van_aarle_fast_2016,
	title = {Fast and flexible {X}-ray tomography using the {ASTRA} toolbox},
	volume = {24},
	copyright = {© 2016 Optical Society of America},
	issn = {1094-4087},
	url = {https://opg.optica.org/oe/abstract.cfm?uri=oe-24-22-25129},
	doi = {10.1364/OE.24.025129},
	language = {EN},
	number = {22},
	urldate = {2025-04-28},
	journal = {Optics Express},
	author = {Aarle, Wim van and Palenstijn, Willem Jan and Cant, Jeroen and Janssens, Eline and Bleichrodt, Folkert and Dabravolski, Andrei and Beenhouwer, Jan De and Batenburg, K. Joost and Sijbers, Jan},
	month = oct,
	year = {2016},
	note = {Publisher: Optica Publishing Group},
	pages = {25129--25147},
}

@article{palenstijn_astra_2011,
	title = {Performance improvements for iterative electron tomography reconstruction using graphics processing units ({GPUs})},
	volume = {176},
	issn = {1047-8477},
	url = {https://www.sciencedirect.com/science/article/pii/S1047847711002267},
	doi = {10.1016/j.jsb.2011.07.017},
	number = {2},
	urldate = {2025-04-28},
	journal = {Journal of Structural Biology},
	author = {Palenstijn, W. J. and Batenburg, K. J. and Sijbers, J.},
	month = nov,
	year = {2011},
	pages = {250--253},
}

@article{han_framing_2018,
	title = {Framing {U}-{Net} via {Deep} {Convolutional} {Framelets}: {Application} to {Sparse}-{View} {CT}},
	volume = {37},
	issn = {1558-254X},
	shorttitle = {Framing {U}-{Net} via {Deep} {Convolutional} {Framelets}},
	url = {https://ieeexplore.ieee.org/document/8332969},
	doi = {10.1109/TMI.2018.2823768},
	number = {6},
	urldate = {2025-04-28},
	journal = {IEEE Transactions on Medical Imaging},
	author = {Han, Yoseob and Ye, Jong Chul},
	month = jun,
	year = {2018},
	pages = {1418--1429},
}

@inproceedings{ronneberger_u-net_2015,
	address = {Cham},
	title = {U-{Net}: {Convolutional} {Networks} for {Biomedical} {Image} {Segmentation}},
	isbn = {978-3-319-24574-4},
	shorttitle = {U-{Net}},
	doi = {10.1007/978-3-319-24574-4_28},
	language = {en},
	booktitle = {Medical {Image} {Computing} and {Computer}-{Assisted} {Intervention} – {MICCAI} 2015},
	publisher = {Springer International Publishing},
	author = {Ronneberger, Olaf and Fischer, Philipp and Brox, Thomas},
	editor = {Navab, Nassir and Hornegger, Joachim and Wells, William M. and Frangi, Alejandro F.},
	year = {2015},
	pages = {234--241},
}

@misc{keras,
  title={Keras},
  author={Chollet, Fran\c{c}ois and others},
  year={2015},
  howpublished={\url{https://keras.io}},
}

@misc{tensorflow,
title={ {TensorFlow}: Large-Scale Machine Learning on Heterogeneous Systems},
url={https://www.tensorflow.org/},
note={Software available from tensorflow.org},
author={
    Mart\'{i}n~Abadi and
    Ashish~Agarwal and
    Paul~Barham and
    Eugene~Brevdo and
    Zhifeng~Chen and
    Craig~Citro and
    Greg~S.~Corrado and
    Andy~Davis and
    Jeffrey~Dean and
    Matthieu~Devin and
    Sanjay~Ghemawat and
    Ian~Goodfellow and
    Andrew~Harp and
    Geoffrey~Irving and
    Michael~Isard and
    Yangqing Jia and
    Rafal~Jozefowicz and
    Lukasz~Kaiser and
    Manjunath~Kudlur and
    Josh~Levenberg and
    Dandelion~Man\'{e} and
    Rajat~Monga and
    Sherry~Moore and
    Derek~Murray and
    Chris~Olah and
    Mike~Schuster and
    Jonathon~Shlens and
    Benoit~Steiner and
    Ilya~Sutskever and
    Kunal~Talwar and
    Paul~Tucker and
    Vincent~Vanhoucke and
    Vijay~Vasudevan and
    Fernanda~Vi\'{e}gas and
    Oriol~Vinyals and
    Pete~Warden and
    Martin~Wattenberg and
    Martin~Wicke and
    Yuan~Yu and
    Xiaoqiang~Zheng},
  year={2015},
}

@article{scikit-image,
	title = {scikit-image: image processing in Python},
	volume = {2},
	issn = {2167-8359},
	url = {https://doi.org/10.7717/peerj.453},
	doi = {10.7717/peerj.453},
	pages = {e453},
	journal= {{PeerJ}},
	author = {van der Walt, Stéfan and Schönberger, Johannes L. and Nunez-Iglesias, Juan and Boulogne, François and Warner, Joshua D. and Yager, Neil and Gouillart, Emmanuelle and Yu, Tony and contributors, the scikit-image},
	date = {2014-06},
}

\renewcommand{\thefigure}{S\arabic{figure}}
\setcounter{figure}{0}  
\renewcommand{\thetable}{S\arabic{table}}
\setcounter{table}{0}

\section*{Supplementary Materials}

\begin{figure*}[h!]
\centerline{\includegraphics[width=\textwidth, trim={0 11cm 0 11cm}, clip]{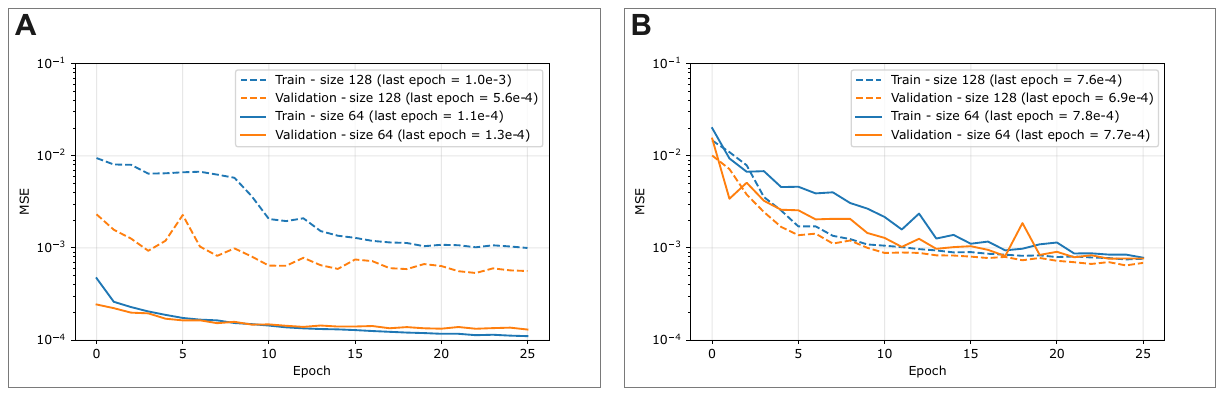}}
\captionof{figure}{\small The mean squared error (MSE) loss curves for 2.5D data reconstructed with 128 views for parallel beam geometry, with a block size of 128 in comparison to a block size of 64 in data preprocessed to \textbf{A) }a wider window setting of (width = 2048, level = 0) HU, and \textbf{B) }a narrower window setting of (width = 1700, level = -600) HU.}\label{fig:loss_window_blockSize}
\end{figure*}

\begin{figure*}[h!]
\centerline{\includegraphics[width=\textwidth, trim={0 11cm 0 11cm}, clip]{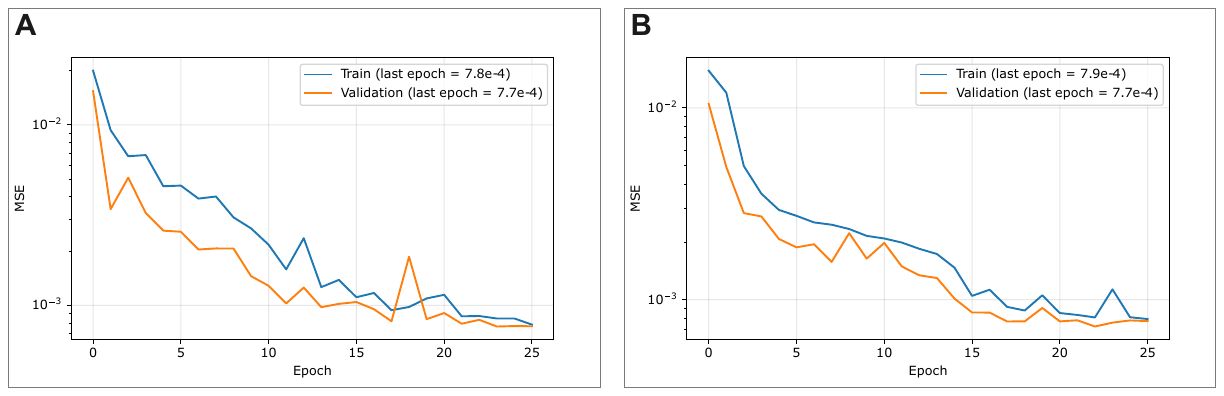}}
\captionof{figure}{\small The mean squared error (MSE) loss curves for data reconstructed with 128 views for parallel beam geometry, formatted to 2.5D with a block size of 64 preprocessed to a narrower window setting of (width = 1700, level = -600) HU, trained with \textbf{A) }the vanilla U-Net, and \textbf{B) }the dual-frame U-Net as reported by \cite{ries_improving_2024}.}\label{fig:loss_unet_comparison}
\end{figure*}

\begin{table*}[h!]
\centering
\captionof{table}{\small Mean MSE and SSIM for Parallel, Fan, and Cone Beam Geometries\\}
\label{tab:geomComp_MSE_SSIM_supp}
  \small
\resizebox{\textwidth}{!}{%
   \begin{tabular}{lc|cccc|cccc|cccc}
    \toprule
      & Views &  \multicolumn{4}{c|}{Parallel} & \multicolumn{4}{c|}{Fan} & \multicolumn{4}{c}{Cone} \\
     \midrule
     MSE $\downarrow$ & & Sparse & 2D & 2.5D & 3D& Sparse& 2D  & 2.5D & 3D& Sparse &2D  & 2.5D & 3D \\
    \midrule
    \midrule
    
    & 32 & $7.21\cdot10^{-3}$& $5.15\cdot10^{-4}$ & $8.62\cdot10^{-4}$ &$5.50\cdot10^{-4}$ & $1.69\cdot10^{-2}$& $8.62\cdot10^{-4}$ & $1.59\cdot10^{-3}$ &$1.25\cdot10^{-3}$ & 
    $2.12\cdot10^{-2}$&
    $1.52\cdot10^{-3}$ & $2.17\cdot10^{-3}$ &$1.75\cdot10^{-3}$\\
    & 64 & $4.79\cdot10^{-3}$& $1.75\cdot10^{-4}$ & $2.64\cdot10^{-4}$ & $1.75\cdot10^{-4}$ &
    $1.17\cdot10^{-2}$
    & $3.37\cdot10^{-4}$ & $5.27\cdot10^{-4}$ & $3.50\cdot10^{-4}$ & 
    $1.51\cdot10^{-2}$
    &$9.68\cdot10^{-4}$ & $1.15\cdot10^{-3}$ & $1.10\cdot10^{-3}$ \\
    & 128 & $3.40\cdot10^{-3}$&$5.99\cdot10^{-5}$ & $8.11\cdot10^{-5}$ & $5.96\cdot10^{-5}$&
    $8.57\cdot10^{-3}$
    &$1.36\cdot10^{-4}$ & $1.78\cdot10^{-4}$ & $1.18\cdot10^{-4}$& 
    $1.17\cdot10^{-2}$
    &$7.29\cdot10^{-4}$ & $6.85\cdot10^{-4}$ & $3.22\cdot10^{-4}$ \\
    
    \midrule
    \midrule
    SSIM $\uparrow$& & Sparse & 2D & 2.5D & 3D& Sparse& 2D  & 2.5D & 3D& Sparse &2D  & 2.5D & 3D \\
    \midrule
    \midrule
    & 32 & 0.343 & 0.884 & 0.840 & 0.872 &0.260
    & 0.859 & 0.783 & 0.795&
    0.323
    & 0.814 & 0.731 & 0.792\\
    & 64 & 0.428 &0.932 & 0.902 & 0.926 & 0.311
     & 0.897 & 0.865 & 0.893 &
     0.381
     & 0.859 & 0.804 & 0.828\\
    & 128 & 0.540 & 0.964 & 0.952 & 0.963 & 0.376 & 0.943 & 0.922 & 0.946 & 
    0.450 &
    0.891 & 0.860 & 0.889\\
    
    \bottomrule
    \end{tabular}}
\end{table*}

\begin{table*}[h!]
\centering
\captionof{table}{\small Comparison of 2D and 2D3ch Mean MSE and SSIM for Parallel, Fan, and Cone Beam Geometries for the Axial 128 Views}
\label{tab:geomComp_MSE_SSIM_128}
  \small
\resizebox{\textwidth}{!}{%
   \begin{tabular}{l|cccc|cccc|cccc}
    \toprule
      128 Views & \multicolumn{4}{c|}{Parallel} & \multicolumn{4}{c|}{Fan} & \multicolumn{4}{c}{Cone} \\
     \midrule
     
       & 2D & 2D3ch & 2.5D & 3D& 2D & 2D3ch & 2.5D & 3D& 2D & 2D3ch & 2.5D & 3D \\
    \midrule
    \midrule
     MSE $\downarrow$ & $5.99\cdot10^{-5}$ & $5.27\cdot10^{-4}$ & $8.11\cdot10^{-5}$ & $5.96\cdot10^{-5}$& $1.36\cdot10^{-4}$ &$1.80\cdot10^{-4}$& $1.78\cdot10^{-4}$ & $1.18\cdot10^{-4}$& $7.29\cdot10^{-4}$ & $2.35\cdot10^{-3}$& $6.85\cdot10^{-4}$ & $3.22\cdot10^{-4}$ \\
    \midrule
     SSIM $\uparrow$& 0.964 & 0.813 & 0.952 & 0.963 & 0.943 & 0.925 & 0.922 & 0.946 & 0.891 & 0.680 & 0.860 & 0.889\\
    
    \bottomrule
    \end{tabular}}
\end{table*}


\end{document}